\documentclass[10pt, a4paper]{article}

\usepackage{lrec-coling2024} 

\usepackage{lscape}
\usepackage{graphicx}
\usepackage{array, calc}
\usepackage{float}
\usepackage{booktabs}
\usepackage{caption} 
\usepackage{amsmath}
\usepackage{comment}
\usepackage{rotating}
\usepackage{natbib}
\usepackage{etoolbox}
\usepackage{appendix}
\usepackage{hyperref}
\usepackage[capitalize]{cleveref}

\crefname{figure}{Figure}{Figures}
\crefname{table}{Table}{Tables}

\AtBeginDocument{%
  \providecommand\BibTeX{{%
    \normalfont B\kern-0.5em{\scshape i\kern-0.25em b}\kern-0.8em\TeX}}}

\title{How Important is Domain Specificity in Language Models and Instruction Finetuning for Biomedical Relation Extraction?}

\name{Aviv Brokman, Ramakanth Kavuluru} 
\address{University of Kentucky, Lexington, KY USA \\
         avivbrokman@gmail.com, ramakanth.kavuluru@uky.edu \\}

\abstract{Cutting edge techniques developed in the general NLP domain are often subsequently applied to the high-value, data-rich biomedical domain. The past few years have seen generative language models (LMs), instruction finetuning, and few-shot learning become foci of NLP research. As such, generative LMs pretrained on biomedical corpora have proliferated and biomedical instruction finetuning has been attempted as well, all with the hope that domain specificity improves performance on downstream tasks. Given the nontrivial effort in training such models, we investigate what, if any, benefits they have in the key biomedical NLP task of relation extraction. Specifically, we address two questions: (1) Do LMs trained on biomedical corpora outperform those trained on general domain corpora? (2) Do models instruction finetuned on biomedical datasets outperform those finetuned on assorted datasets or those simply pretrained? We tackle these questions using existing LMs, testing across four datasets.  In a surprising result, general-domain models typically outperformed biomedical-domain models. However, biomedical instruction finetuning improved performance to a similar degree as general instruction finetuning, despite having orders of magnitude fewer instructions. Our findings suggest it may be more fruitful to focus research effort on larger-scale biomedical instruction finetuning of general LMs over building domain-specific biomedical LMs.
 \\ \newline  }

\begin{document}

\maketitleabstract

\section{Introduction}
Biomedical entities and relations among them are at the forefront of biomedical knowledge discovery. Novel relational information among biomedical entities is often conveyed through text in scientific literature or clinical notes. For instance, protein-protein interactions (to understand disease etiology and progression), gene-disease associations (to identify potential drug targets), drug-disease treatment relations (to spot off-label usage or assess potential for repositioning), and drug-gene interactions (to design targeted therapies) are often discussed in literature. The extracted relations are often used to build knowledge graphs to facilitate knowledge discovery and knowledge-based search systems~\citep{bakal2018exploiting,lever2018collaborative,zhang2021drug}. Adverse event relations (linking medications and adverse effects) are often reported in clinical text and their extraction  facilitates more effective post-market pharmacovigilance~\citep{liu2019towards,botsis2012vaccine}. As such, biomedical relation extraction (RE) is a frequently pursued task in the BioNLP community, featuring in many shared task series (e.g., BioCreative, i2b2, n2c2). 
Our current contribution is about high-level questions regarding the value of biomedical domain specific language models (LMs) and instruction finetuning for biomedical RE. In the rest of this section, we overview the recent RE relevant methods landscape and frame our contributions in that context.

The last few years have seen a proliferation of  transformer-based generative (encoder-decoder and decoder-only) LMs. In part, these have become popular because tasks such as summarization, paraphrasing, and code generation  are fundamentally generative.  But generative models have advantages over encoder-only LMs even in some tasks that were previously addressed with the latter.  Encoder models were largely designed for classification problems, where a classifier is trained on top of the final layer embedding of the [CLS] token.  They have been used in myriad other ways --- notably in named entity recognition (NER) and RE where embeddings of spans are fed to classifiers \citep{spert, jerex, pure}.  These methods finetune weights from scratch and therefore require ample data.  But few- and zero-shot learning have become focal areas of NLP research; their maturation promises improved scalability of NLP, as manual creation of large datasets is time consuming.  Few-/zero-shot learning in encoder models is often achieved by aligning the NLP task objective function with the pretraining objective function. Prompting with templates containing [MASK] tokens to be predicted has been the few-/zero-shot method for encoder-only models \citep{schick-schutze-2021-exploiting, wang-etal-2022-automatic}.  Using such a strategy largely rests on whether the desired NLP task is a classification task or can be formulated as one.  In summation, encoder-only LMs are well-suited to \textit{full finetuning} (as opposed to the few-shot setting) for a variety of non-generative tasks.
The additional  layering and customization required on top of encoder models for RE can be obviated in generative models, by contrast, given their relatively flexible approach of formulating RE with straightforward natural language templates.  This particularly comes in handy in few- and zero-shot learning.

In the last two years, instruction finetuning (IFT) in the general-domain has become a hot area of NLP research for generative LMs \citep{instruct-gpt, wei-ift, sanh-ift, instruct-gpt, flan, llama2}.  In its basic form, IFT is conducted by converting examples of a wide variety of labeled datasets into pairs of natural language instructions (along with input text) and responses (containing the correct output), and then training a generative LM on them.  This typically improves few-shot performance by aligning the model towards the goals that human users have.  Then, for a new task, the corresponding instructions are generally more in-distribution than they would have been sans IFT.  

Most advances in NLP are originally developed for the general domain, and only later adapted to the biomedical domain.  Recently, this has manifested in the development of biomedical versions of popular general-domain LMs.  Intuitively, it is sensible to do so because the distribution of text in biomedical NLP tasks is more similar to biomedical text than to general-domain text, and because biomedical-specific tokenizers can more parsimoniously represent biomedical entities. Likewise, inspired by the striking success of IFT, \citet{inboxbart} instruction finetuned BART-Base \citep{bart} on a biomedical meta-dataset they assembled (called BoX) to create In-BoXBART.  

Biomedical generative LMs are quickly being trained, and we can safely assume that larger biomedical models will be trained in the near-future.  Given the high labor, monetary, and effort costs of creating biomedical-specific LMs, it is prudent to ask if they are worthwhile.  To some degree, this has already been tested in biomedical versions of T5 \citep{t5, scifive} and BART \citep{bart, biobart} in the context of full-finetuning, yielding underwhelming improvements in scores.  However, the tasks tested largely require extremely short, simple generations. Plausibly, these models would differentiate themselves from their general-domain counterparts on more complex tasks, especially in the few-shot context. We investigate whether this is the case in the high-value task of RE\footnote{Here we are explicitly dealing with end-to-end RE, where entities must be extracted by the RE system and are not provided \textit{a priori}.}.  Specifically we ask:

\begin{itemize}
    \item Do generative models pretrained on PubMed-based scientific text outperform those trained on general-domain text?
    \item How do biomedical and general-domain instruction finetuned models perform compared to their base models?
    \item Are the answers to these questions different in the fully finetuned and few-shot settings?
\end{itemize}

We test these questions across four biomedical RE datasets using a variety of open biomedical and general domain LMs. (We make our code available for review\footnote{Code available at \url{https://anonymous.4open.science/r/domain_specificity-BCB8}} and intend to make it public if accepted.)

\section{Related Work}
\subsection{Generative Relation Extraction}
RE has traditionally been conducted via a pipeline in two steps --- NER followed by relation classification (RC) between  pairs
of entities from NER. However, this pipeline approach was joined by end-to-end approaches that employ a joint loss function for both NER and RC phases beginning with \citet{miwa2014modeling}.  Among them, \citet{zeng-etal-2018-extracting} used a copy mechanism to tackle generative RE without an intermediate NER step; several extensions followed suit \citep{nayak2020effective, hou2021discovering, nayak2020effective, zeng2020copymtl, seq2rel}.  One advantage of their strategy is that generation can naturally handle discontinuous entities (e.g., \emph{hand pain} in the phrase \emph{hand and arm pain}). Discontinuous entities pose such a challenge to NER that there is an entire line of research dedicated to solving it \citep{wang-etal-2021-discontinuous, dirkson-etal-2021-fuzzybio}. But these copy-based models require training neural network components from scratch, and are therefore not well-suited to few-shot learning.

\citet{rebel} use BART to directly generate relation triples using special tokens to demarcate roles in the triple; this still requires training randomly initialized special tokens.  \citet{biogpt} finetune BioGPT on natural language templates, aligning the current NLP task's output form most closely with the pretraining strategy.  At test time, relations are extracted from generated text using regular expressions.  We adopt this general strategy of \citet{biogpt} in the present study.  \citet{wadhwa2023revisiting} finetune Flan-T5 \citep{flan} using templates that convert relations into a target sequence of structured data, rather than natural language.  We note that in preliminary experiments, our performance was much lower with templates of this form.

\subsection{Biomedical Language Models}
Early transformer-based biomedical LMs had encoder-only architectures, usually based on BERT \citep{bert}, and were trained on some combination of PubMed abstracts, PubMed Central full-text articles (PMC), and MIMIC III \citep{mimic} clinical notes. BioBERT \citep{biobert} was initialized using weights from BERT and then continually pretrained on PubMed and PMC. \citet{clinicalbert} continued pretraining BioBERT on the MIMIC-III corpus, yielding ClinicalBERT.  Likewise, \citet{biomed-roberta} continually pretrained RoBERTa \citep{roberta} on additional full-text biomedical articles from S2ORC \citep{s2orc}, obtaining BioMed-RoBERTa. Models that trained from scratch on biomedical text generally perform better than their continually trained counterparts.  The BERT-based BlueBERT \citep{bluebert} and PubMedBERT \citep{pubmedbert}, and the RoBERTa-based Bio-LM \citep{bio-lm} are a few such models, with PubMedBERT being a popular base model for many downstream applications. \citet{biomegatron} trained versions of BioMegatron, both from scratch and through continued pre-training from Megatron~\citep{megatron}. BioELECTRA \citep{bioelectra}, a biomedical version of the high-performing ELECTRA \citep{electra}, was trained from scratch on PubMed and PMC. 

As generative models (with encoder-decoder and decoder-only architectures) have risen in popularity, PubMed-based generative models have been trained from scratch as well.  Biomedical versions of T5 \citep{t5} --- SciFive, \citep{scifive} and GPT --- BioGPT \citep{biogpt} and BioMedLM \citep{biomedlm} have all been trained from scratch on PubMed/PMC corpora. BioBART \citep{biobart} is a continually pretrained BART \citep{bart} model with PubMed text.  SciFive and BioBART showed negligible improvement on most biomedical tasks, and modest improvements in a small number of tasks, over their general-domain counterparts. However, we believe the generative capabilities of SciFive and BioBART are under-explored by \citet{scifive} and \citet{biobart} because either the tasks tackled require modest generation, (e.g., multiple choice answer) or are evaluated by ROUGE scores, which only provide rough estimates of generation quality compared to human evaluations.  \citet{biogpt}  tested BioGPT on three end-to-end RE datasets, where it substantially outperformed the comparably-sized GPT-2 Medium \citep{gpt2}.  We are unable to find any performance scores for BioGPT-Large on end-to-end RE.  BioMedLM \citep{biomedlm} scored much higher than the comparably-sized GPT-Neo 2.7B model on three QA tasks.  

In summary, there is evidence that on very short generation tasks, in a fully finetuned setup,  biomedical  encoder-decoder LMs outperform general-domain counterparts.  But there is a dearth of information about the efficacy of decoder-only model performance in more complex generation tasks, and even more so for encoder-decoder models.  Few-shot performance patterns of performance are especially unknown.

\subsection{Instruction Finetuning}
\citet{wei-ift} explore the effect of IFT on zero-shot performance by instruction finetuning on 60 tasks, constructing 10 templates per task, finding consistent, strong improvements in performance.  \citet{sanh-ift} conduct similar experiments on a different set of datasets, yielding the T0 model. \citet{chung2022scaling} build on these previous attempts, ramping up the scale of finetuning up to 1600 tasks and consequently release the Flan-T5 series of instruction finetuned models. InstructGPT \citep{instruct-gpt} is trained using reinforcement learning with human feedback (RLHF) \citep{rlhf-christiano, rlhf-stiennon} on instructions to align GPT-3 \citep{gpt3} to human preferences.  \citet{self-instruct} create synthetic instruction data for IFT using GPT-3, attaining competitive performance with the more expensively trained InstructGPT. In-BoXBART \citep{inboxbart}, the only attempt at biomedical instruction finetuning (to our knowledge), is an IFT version of BART-Base trained on 32 biomedical tasks, with one template per task, making its scale of IFT considerably smaller than that of other IFT models.  

\section{\label{sec:task-setup} Task Setup}
\subsection{Training}

\begin{figure*}
\begin{center}
\includegraphics[scale = 1]{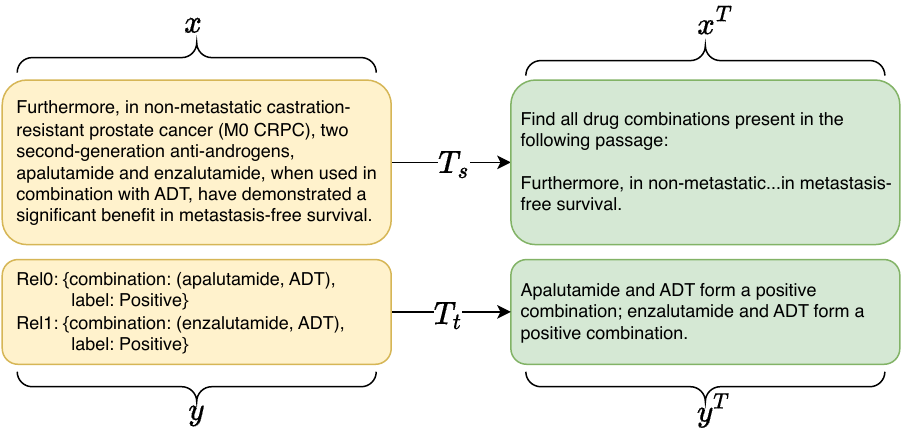}
\caption{\label{fig:template} Conversion of training example of the DCE dataset into natural language sequences.}
\end{center}
\end{figure*}

RE instances generally consist of a source text and a set of semi-structured relations that are asserted in it.  To finetune generative LMs on a relation extraction dataset, we convert its training examples into sequences of natural language before finetuning the models in the usual manner. In essence, the natural language sequences contain the source text along with an instruction to extract relations followed by the relations expressed in natural language form. \cref{fig:template} illustrates our approach  for the drug combination extraction (DCE) task of extracting combination drug therapies from scientific text \citep{n-ary-drug-combos} as we formalize below. 

Let $\left(x,y\right)$ be a training example, where $x$ is the input text and $y$ denotes the relations present present in it.  To finetune LMs, we construct a single prompting template $T = (T_s, T_t)$ for each dataset.  A template consists of a source transformation function $T_s$ and a target transformation function $T_t$, which return source sequence $x^T$, containing the source text $x$ along with an instruction to guide generation, and target sequence $y^T$, conveying the relations $y$ in the form of natural language:
\begin{align*}
T_{s}\left(x\right) &= x^{T}   \\
T_{t}\left(y\right)  &=  y^{T} 
\end{align*} 

For encoder-decoder models, $x^T$ and $y^T$ respectively are supplied to the encoder and decoder modules.  For decoder models, the sequences $x^T$ and $y^T$ are joined into one sequence with a function $J$, which then serves as a single input to the model.  We take $J$ to be 
\[
J\left( x^T, y^T \right) = x^T\|\text{\textbackslash n\textbackslash n}\|y^T 
\]
where $\|$ represents concatenation.  Our formulation of $J$ separates the source and target sequences with two line breaks, though other formulations can be used.  


\subsection{Evaluation}
For validation and test examples, we only have source sequences, and our aim is to generate the target sequences. For encoder-decoder models,  $x^T$ again serves as input to the encoder.  For decoder models, the input sequence is
\[
J\left( x^T,\cdot  \right) = x^T\|\text{\textbackslash n\textbackslash n}.
\]
The generated output sequence of the model  $\hat{y}^T$ must be transformed into a structured format so that predicted relations can be compared to annotated relations to calculate performance metrics.  We design an information extractor for each template using regular expressions.

\section{\label{sec:experimental-setup} Experimental Setup}
We conduct two experiments to investigate the effect of domain-specificity. First, we compare biomedical- and general-domain models. Second, we compare biomedical instruction fine-tuned, general-domain instruction fine-tuned, and base model performance. All experiments are repeated in fully finetuned and few-shot settings.  \textit{We will include appendices with templates for each dataset as well as modeling choices (learning rate, batch size, \ldots)  in the final camera-ready version (if accepted), given conference submission instructions restrict their inclusion during initial review.}

\subsection{\label{subsec:experiments} Experiments}
\subsubsection{\label{subsubsec:base-models} Biomedical vs General-Domain Pretraining}
In our first experiment, we compare the efficacy of biomedical LMs and their closest general-domain language equivalent in terms of architecture and size. We evaluate performance of the biomedical decoder LMs BioGPT (350M) \citep{biogpt}, BioGPT-Large (1.5B) \citep{biogpt}, and BioMedLM (2.7B) \citep{biomedlm} and compare to their approximate general domain equivalents, GPT2-Medium (350M), GPT2-XL (1.5B) \citep{gpt2}, and GPT-Neo (2.7B) \citep{gpt-neo}. We also evaluate performance of the biomedical encoder-decoder LMs SciFive-Base (220M), SciFive-Large (770M) (both PubMed version) \citep{scifive}, BioBART-Base (140M), and BioBART-Large (400M) \citep{biobart} and compare to the general domain equivalents T5-Base (220M), T5-Large (770M) \citep{t5}, BART-Base (140M), and BART-Large (400M) \citep{bart}.

\subsubsection{\label{subsubsec:old-ift} Instruction Finetuning vs Base Models}
In our second experiment, we compare the efficacy of instruction finetuned models and their base models.  No biomedical models have been instruction finetuned in our literature review.  General-domain models have been instruction finetuned with general-domain datasets and separately with biomedical datasets.  For general-domain instructions, we compare the Flan-T5 \citep{flan} series of models to the base T5 models \citep{t5} they were trained from.  For biomedical domain instructions, we compare In-BoXBART \citep{inboxbart} to the base BART \citep{bart}, from which it was trained. 

\subsubsection{Few-Shot Setting}
We repeat the experiments in Sections \ref{subsubsec:base-models} and \ref{subsubsec:old-ift}, but in the few-shot setting.  We emphasize that we finetune the models, rather than perform in-context learning.  We experiment with 16-shot and 64-shot learning.  Few-shot performance values are averaged over five runs with different random seeds for selecting training examples.  Barring a single exception, fully finetuned decoder-only model performances were quite inferior (F1 $<$ 30) compared to encoder-decoder models; so we did not perform few-shot experiments on decoder-only models.

\subsection{\label{subsec:data} Datasets}
We use four public biomedical  RE datasets for our experiments.  For datasets without an established train/validation split, we select 20\% of the training dataset to serve as a validation set.

\subsubsection{\label{subsubsec:CDR} CDR}
CDR \citep{bc5cdr} is a BioCreative V dataset consisting of PubMed abstracts for which all mentions of chemicals and diseases are annotated, as well as all relations between them describing chemical-induced diseases.  Relations all belong to a single class and are annotated at the entity level. 1500 abstracts are split into train, validation, and test sets of equal size.

\subsubsection{\label{subsubsec:ChemProt} ChemProt }
ChemProt \citep{chemprot} is a BioCreative VI dataset consisting of PubMed abstracts for which all mentions of chemical compounds/drugs and genes/proteins are annotated, as well as all intra-sentence relations between the two entity types belonging to a set of relation types of interest.  Relations belong to the classes \textit{agonist}, \textit{antagonist}, \textit{upregulate}, \textit{downregulate}, and \textit{substrate}, and are annotated at the mention level.  Since relations are annotated at the mention level, an example may contain multiple annotated relations that are conceptually identical but have different mentions.  The dataset is split into 1,020 training, 612 validation, and 800 test examples.

\subsubsection{\label{subsubsec:n-ary-drug} DCE}
The Drug Combinations Extraction (DCE) \citep{n-ary-drug-combos} dataset consists of sentences selected from PubMed abstracts containing multiple drugs, annotated for drugs given in combination.  Sets of drugs are annotated as consisting of a drug combination with a positive effect, a drug combination with an effect not clarified by the text, or not a drug combination. Though some studies combine the latter two relation types, \citep{n-ary-drug-combos, yuhang-drug-combos} we model them as-is. The dataset is split into 1362 training and 272 test instances.  

\subsubsection{\label{subsubsec:DDI} DDI}
The DDI \citep{ddi} dataset consists of texts from DrugBank and Medline annotated for drugs/chemicals and relations holding for pairs of co-occurring drugs.  Mentions of drugs are classified as generic drug name, brand-name drug, drug group, or non-drug active substance.  Relations holding for drug pairs are \textit{pharmacokinetic mechanism}, \textit{combination effect}, \textit{advice regarding interaction}, or \textit{interaction in the absence of supporting information}.  Relations are annotated at the mention level.  The dataset consists of 784 DrugBank documents and 233 Medline abstracts, split up into 714 train and 303 test examples.

\subsection{Evaluation}
We calculate precision, recall, and F1 for all tasks.  Evaluation is done in a strict setting: relations are only considered correct if the entities are all correct and relation type, if applicable, is correct.  A correct entity consists of an exact string match.  For CDR, the predicted entity must match one of the mentions corresponding to the MeSH ID of the gold entity.

\section{Results}

\begin{figure*}
\begin{center}
\includegraphics[scale = 1]{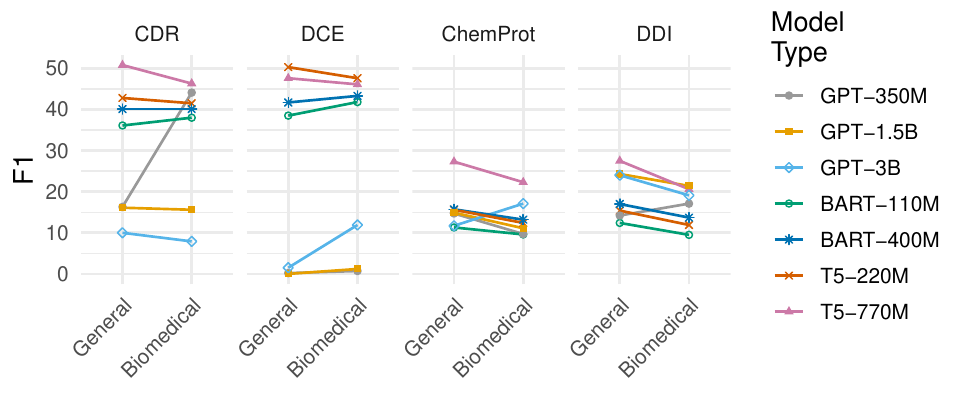}
\caption{\label{fig:pretraining-full-shot} Comparison of fully finetuned LMs pretrained on general-domain and biomedical text.}
\end{center}
\end{figure*}

\begin{figure*}
\begin{center}
\includegraphics[scale = 1]{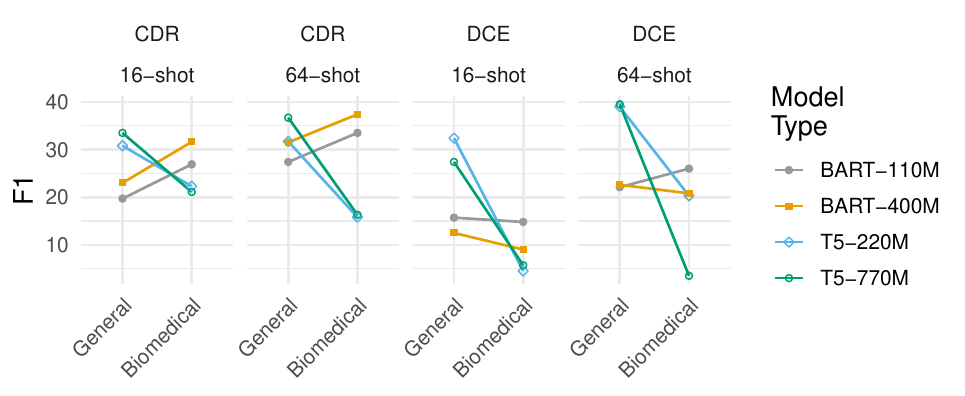}
\caption{\label{fig:pretraining-few-shot} Comparison of few-shot finetuned LMs pretrained on general-domain and biomedical text.}
\end{center}
\end{figure*}

\begin{figure*}
\begin{center}
\includegraphics[scale = 1]{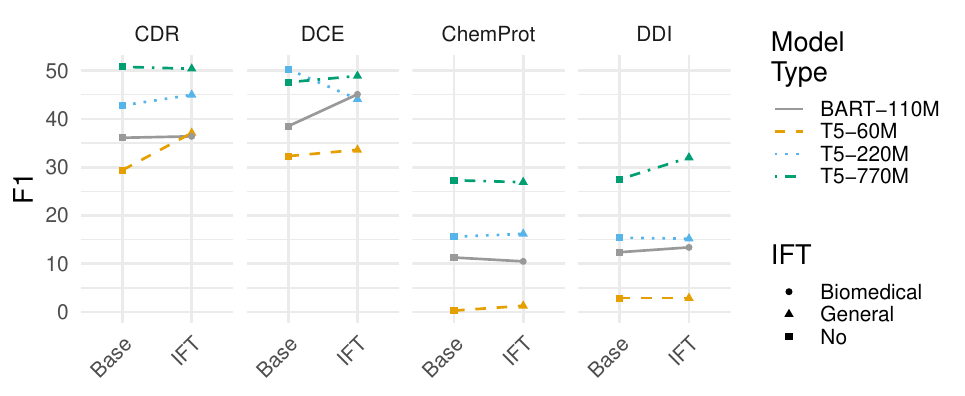}
\caption{\label{fig:ift-full-shot} Comparison of fully finetuned base and instruction-finetuned LMs across all datasets.}
\end{center}
\end{figure*}

\begin{figure*}
\begin{center}
\includegraphics[scale = 1]{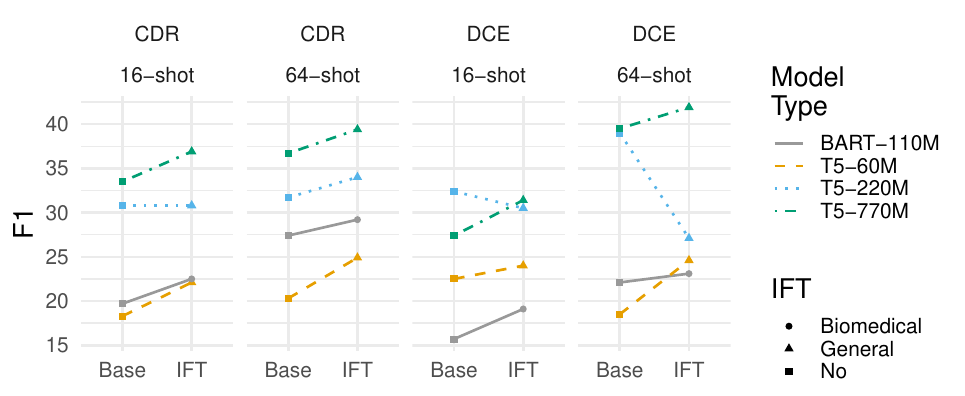}
\caption{\label{fig:ift-few-shot} 
Few-shot comparison of   base and instruction finetuned LMs on high-performing datasets.}
\end{center}
\end{figure*}

All results are displayed in Table~\ref{table:master} (page 8), primarily grouped into model series (e.g., T5) with an in-group ordering by model size. Before moving into the main results, we note two high level  trends   among the datasets and model series: (1) Performance was higher in general for CDR and DCE than in ChemProt and DDI, perhaps owing to the higher number of relation types in the latter datasets.  (Henceforth, we refer to CDR and DCE as high-performing and ChemProt and DDI as low-performing as we explain other results). (2) Encoder-decoder models performed much better than decoder-only models, with the exception of BioGPT for CDR.  In fact, decoder-only models performed so poorly that we  largely limit our further discussion to encoder-decoder models. Potentially, this is because decoder  models may be more amendable to generative outputs such as summaries and not as suitable for RE. 

\begin{table*}
\renewcommand{\arraystretch}{1.2}
\begin{center}
\begin{tabular}{@{}lcrrrrrrrr@{}}
\toprule
                     &         & \multicolumn{3}{c}{CDR} & \multicolumn{3}{c}{DCE} & ChemProt & DDI  \\ \cmidrule(lr){3-5} \cmidrule(lr){6-8} \cmidrule(lr){9-9} \cmidrule(l){10-10}    
Model                & \# Par. & 16   & 64   & Full & 16   & 64   & Full & Full & Full \\ \midrule
GPT-2-Medium         & 350M    & -    & -    & 16.3 & -    & -    & 0.2  & 14.7 & 14.2 \\
BioGPT\textsuperscript{b}               & 350M    & -    & -    & 44.1 & -    & -    & 0.7  & 9.7  & 17.1 \\
GPT-2-XL             & 1.5B    & -    & -    & 16.1 & -    & -    & 0    & 14.9 & 24.3 \\
BioGPT-Large\textsuperscript{b}         & 1.5B    & -    & -    & 15.6 & -    & -    & 1.2  & 11.1 & 21.4 \\
GPT-Neo-2.7B         & 2.7B    & -    & -    & 10.0 & -    & -    & 1.5  & 11.7 & 24.0 \\ 
BioMedLM\textsuperscript{b}             & 2.7B    & -    & -    & 7.9  & -    & -    & 11.9 & 17.1 & 19.1 \\ \midrule 
BART-Base            & 110M    & 17.6 & 27.4 & 36.1 & 8.9  & 30.6 & 38.5 & 11.3 & 12.4 \\
BioBART-Base\textsuperscript{b}         & 110M    & 29.0 & 33.1 & 38.0 & 8.9  & 24.1 & 41.8 & 9.6  & 9.5  \\
In-BoXBART\textsuperscript{$\dagger$}           & 110M    & 21.4 & 28.9 & 36.4 & 10.4 & 26.7 & 45.1 & 10.5 & 13.4 \\
BART-Large           & 400M    & 26.4 & 32.5 & 40.1 & 9.6  & 30.4 & 41.7 & 15.7 & 17.0 \\
BioBART-Large\textsuperscript{b}        & 400M    & 30.7 & 33.0 & 40.1 & 8.7  & 29.2 & 43.3 & 13.2 & 13.7 \\ \midrule
T5-Small             & 60M     & 18.3 & 20.3 & 29.4 & 22.5 & 18.5 & 32.3 & 0.3  & 2.9  \\
Flan-T5-Small*        & 60M     & 18.9 & 22.2 & 37.1 & 10.7 & 21.2 & 33.6 & 1.3  & 2.9  \\
T5-Base              & 220M    & 30.3 & 29.8 & 42.8 & 30.0 & 39.3 & 50.3 & 15.6 & 15.4 \\
SciFive-Base\textsuperscript{b}         & 220M    & 28.8 & 10.1 & 41.5 & 0    & 21.6 & 47.6 & 12.4 & 11.9 \\
Flan-T5-Base*         & 220M    & 30.5 & 33.1 & 45.0 & 30.0 & 33.8 & 44.1 & 16.2 & 15.2 \\
T5-Large             & 770M    & 33.9 & 37.7 & 50.8 & 25.7 & 42.4 & 47.6 & 27.3 & 27.5 \\
SciFive-Large\textsuperscript{b}        & 770M    & 22.7 & 5.9  & 46.3 & 3.1  & 3.7  & 46.1 & 22.3 & 20.6 \\
Flan-T5-Large*        & 770M    & 35.9 & 38.4 & 50.4 & 27.4 & 45.1 & 48.9 & 26.9 & 32.0 \\
\bottomrule
\end{tabular}
\caption{\label{table:master} Comparison of model F1-scores  across a range of model types and sizes for four biomedical RE datasets. Few-shot values are averages across five different random seeds for selecting training examples.  Models marked with ``b'' are biomedically pretrained.  Models marked with * are instruction finetuned with general-domain instructions, and those marked by $\dagger$ are instruction finetuned with biomedical instructions.}
\end{center}
\end{table*}

\subsection{Biomedical vs General-Domain Pretraining}
Overall, general-domain pretraining was superior to biomedical-domain pretraining in full and few-shot finetuning experiments (\cref{fig:pretraining-full-shot} and \cref{fig:pretraining-few-shot}). In full finetuning, this pattern held for T5 models across all datasets and for BART models in low-performing datasets.  Notable exceptions are the BART models in high-performing datasets, where biomedical models outperformed, or performed similarly to, corresponding general-domain models, and BioGPT which substantially outperformed its GPT2-Medium counterpart.  However, BioGPT performed poorly in all other circumstances.  In few-shot finetuning, biomedical T5 models performed much worse than general-domain models (\cref{fig:pretraining-few-shot}).  Biomedical pretraining had inconsistent effects in BART models, improving performance for CDR while generally lowering performance for DCE.

\subsection{Effect of Instruction Finetuning }
In full finetuning, IFT versions of models generally performed similarly or slightly better than their base versions (\cref{fig:ift-full-shot}). The small benefits of IFT mostly manifested in high-performing datasets. In few-shot finetuning, both general-domain instructions and biomedical instructions consistently improved performance  and to a similar degree (\cref{fig:ift-few-shot}), even though there were orders of magnitude fewer biomedical instruction datasets.

\section{Discussion}

We were surprised that decoder-only models performed so poorly  and so much worse than encoder-decoder models.  \citet{biogpt} found BioGPT to attain state-of-the-art (SOTA) performance on CDR, DDI, and KD-DTI \citep{kd-dti}.  While we differed from \citet{biogpt} in some preprocessing of the dataset (e.g., we did not remove examples containing no relations, how many epochs we trained for, and how we determined if predicted entities matched gold entities), our performance on DDI was drastically lower than theirs, and our results on two additional datasets, DCE and ChemProt are far below SOTA. An unusual finding is that only for CDR, BioGPT performance is better than BioGPT-Large and BioMedLM (2.7B models).  Preliminary experiments with Llama2-7b \citep{llama2} were similarly fruitless and hence we do not report any scores given disappointing initial experiments.  As encoder-decoder models performed so much better, future practical generative RE efforts may be more fruitful with encoder-decoder architectures; and if additional models are to be pretrained on biomedical corpora (PubMed), encoder-decoder models may be a safer bet.  

However, training additional generative LMs (even encoder-decoder types) on PubMed generally yielded inferior models, at least as far as RE is concerned.  This result held in both full and few-shot finetuning contexts.  We found this pattern counterintuitive, since most of the datasets we experimented with consist of annotated texts from PubMed --- one might expect that models trained on PubMed itself would be particularly well suited for these tasks, as they fall squarely inside of the pretraining corpus distribution.  We hypothesize that (1) the sheer quantity of tokens available for training on general-domain text outweighs the benefits of domain-specificity, and/or (2) that biomedical-domain models may learn inferior linguistic representations as biomedical models are trained on a comparatively narrow subset of the distribution of the English language, and RE requires linguistic sophistication. Thus, even for encoder-decoder models, training biomedical LMs from scratch may not be judicious.

As for the IFT experiments, our results do not suggest that domain-specific training is altogether pointless. In the full finetuning case, IFT models achieved roughly the same performance as their base models, though they converged in far fewer epochs during training. In the few-shot setting, we found that small-scale biomedical IFT produced performance gains comparable to large-scale general-domain IFT.  In-BoXBART was finetuned on 32 biomedical tasks with one template per task. By contrast, Flan-T5 models were finetuned on 1,836 tasks with multiple templates per task, and \citet{flan} found that performance continued to climb until \textasciitilde 300 tasks.  This suggests that IFT with additional biomedical datasets could yield strong benefits.  While there are far fewer annotated biomedical datasets than general-domain datasets, the biomedical metadataset BigBIO \citep{bigbio}, containing over 100 datasets currently, would be a valuable resource for biomedical IFT.  Alternatively, synthetic biomedical instruction examples could also be generated \textit{\`a la} \citet{self-instruct}.
Some nuance is required when interpreting the few-shot performance of In-BoxBART on the CDR dataset.  Among the tasks that In-BoxBART was instruction finetuned on is the NER task for CDR.  It is possible that some of In-BoxBART's capabilities on RE for CDR come from this sort of leakage.
In addition to our previous suggestions for further research, our conclusions are limited to RE, a particularly high-value application of LMs.  Other biomedical tasks should be evaluated as well.

To conclude, the fast evolving landscape of public LMs and their use in many fields naturally leads to domain specific models and IFT. In this paper, we took the first step toward rigorous assessment of the need for domain specificity in these models for biomedical RE. It would be interesting to see if our findings apply to other tasks (e.g., biomedical summarization) and domains (e.g., legal) as well. 


\section{Bibliographical References}\label{sec:reference}
\bibliographystyle{lrec-coling2024-natbib}
\bibliography{domain-specificity}

\end{document}